\RequirePackage{fix-cm}
\documentclass[sn-mathphys]{sn-jnl}                    
\usepackage{graphicx}

\usepackage{amsmath}
\usepackage{algorithm}
\usepackage{epsfig}
\usepackage{subfigure}
\usepackage{graphics}
\usepackage{color,ulem}
\usepackage{amssymb}
\usepackage{algorithm,hyperref}

\begin{document}

\title[Scalable graph neural networks for predictions of HOMO-LUMO gap]{Scalable training of graph convolutional neural networks for fast and accurate predictions of HOMO-LUMO gap in molecules}

\author*[1]{\fnm{Jong Youl} \sur{Choi}}\email{choij@ornl.gov}

\author[2]{\fnm{Pei} \sur{Zhang}}\email{zhangp1@ornl.gov}

\author[1]{\fnm{Kshitij} \sur{Mehta}}\email{mehtakv@ornl.gov}

\author[2]{\fnm{Andrew} \sur{Blanchard}}\email{blanchardae@ornl.gov}

\author[2]{\fnm{Massimiliano} \sur{Lupo Pasini}}\email{lupopasinim@ornl.gov}

\affil*[1]{\orgdiv{Computer Science and Mathematics Division}, \orgname{Oak Ridge National Laboratory}, \orgaddress{\street{1 Bethel Valley Road}, \city{Oak Ridge}, \postcode{37831}, \state{TN}, \country{USA}}}

\affil[2]{\orgdiv{Computational Sciences and Engineering Division}, \orgname{Oak Ridge National Laboratory}, \orgaddress{\street{1 Bethel Valley Road}, \city{Oak Ridge}, \postcode{37831}, \state{TN}, \country{USA}}}


\abstract{
Graph Convolutional Neural Network (GCNN) is a popular class of deep learning (DL) models in material science to predict material properties from the graph representation of molecular structures. 
Training an accurate and comprehensive GCNN surrogate for molecular design requires large-scale graph datasets and is usually a time-consuming process.
Recent advances in GPUs and distributed computing open a path to reduce the computational cost for GCNN training effectively.
However, efficient utilization of high performance computing (HPC) resources for training requires simultaneously optimizing large-scale data management and scalable stochastic batched optimization techniques.

In this work, we focus on building GCNN models on HPC systems to predict material properties of millions of molecules. 
We use HydraGNN, our in-house library for large-scale GCNN training, leveraging \textit{distributed data parallelism} in PyTorch.
We use ADIOS, a high-performance data management framework for efficient storage and reading of large molecular graph data.
We perform parallel training on two open-source large-scale graph datasets 
to build a GCNN predictor for an important quantum property known as the HOMO-LUMO gap.
We measure the scalability, accuracy, and convergence of our approach on two DOE supercomputers: the Summit supercomputer at the Oak Ridge Leadership Computing Facility (OLCF) and the Perlmutter system at the National Energy Research Scientific Computing Center (NERSC).
We present our experimental results with HydraGNN showing i) reduction of data loading time up to 4.2 times compared with a conventional method and ii) linear scaling performance for training up to 1,024 GPUs on both Summit and Perlmutter.
}


\maketitle

\begin{abstract}
This is the abstract

\end{abstract}

{\noindent
Notice: This manuscript has been authored by UT-Battelle, LLC under Contract No. DE-AC05-00OR22725 with the U.S. Department of Energy.  The publisher, by accepting the article for publication, acknowledges that the U.S. Government retains a non-exclusive, paid up, irrevocable, world-wide license to publish or reproduce the published form of the manuscript, or allow others to do so, for U.S. Government purposes. The DOE will provide public access to these results in accordance with the DOE Public Access Plan (\url{http://energy.gov/downloads/doe-public-access-plan}).
}\\

\section{Introduction}\label{sec1}


Drug discovery and molecular design rely heavily on predicting material properties directly from their atomic structure. A particular property of interest for molecular design is the energy gap between the highest occupied molecular orbital (HOMO) and lowest unoccupied molecular orbital (LUMO), known as the HOMO-LUMO gap. The HOMO-LUMO gap is a valid approximation for the lowest excitation energy of a molecule and is used to express its chemical reactivity. In particular, molecules that are more chemically reactive are characterized by a lower HOMO-LUMO gap.  
There are many physics-based computational approaches to compute the HOMO-LUMO gap of a molecule such as \textit{ab initio} molecular dynamics (MD)~\cite{car-parrinello, marx} and density-functional tight-binding (DFTB)~\cite{sokolov2021analytical}.
While these methods have been instrumental in predictive materials science, 
they are extremely computationally expensive. 
The advent of deep learning (DL) models has provided alternative methodologies to produce fast and accurate predictions of material properties and hence enable rapid screening in the large search space to select material candidates with desirable properties~\cite{gaultois, lu, gomez, xue}.

In particular, graph convolutional neural network (GCNN) models are extensively used in material science to predict material properties from atomic information
\cite{cgcnn, megnet}. 
When GCNN models are used as surrogates in screening of the vast chemical space, the models have to process large amounts of streaming data which is dynamically produced by molecular design applications 
\cite{Reymond2015}.

To effectively process large volumes of data in training large complex GCNN models, 
both data loading and model training must scale on multi-node hybrid CPU-GPU high-performance computing (HPC) resources. HPC techniques to scale the training use distributed data parallelism (DDP) to distribute data in batches across different processes. Each process computes gradient updates for the coefficients of the DL model on the local batch, and combines local gradient updates of all processes by averaging them. Although DDP is a well established technique, the specific use of DDP to scale the training of GCNN models is still largely unexplored.

In this work we analyze the scalability of our library of GCNN models on two open-source graph datasets describing the HOMO-LUMO gap for a wide variety of molecules: the PCQM4Mv2 dataset from the Open Graph Benchmark (OGB)~\cite{Hu2020,hu2021ogb} and the AISD HOMO-LUMO dataset~\cite{Blanchard2022AISD} generated at Oak Ridge National Laboratory (ORNL). The scalability of the data loading and GCNN training on the two datasets has been tested on the Summit supercomputer at the Oak Ridge Leadership Computing Facility (OLCF) and the Perlmutter supercomputer at the National Energy Research Scientific Computing Center (NERSC).
For our study, we use HydraGNN, a library we have developed for scalable data reading and GCNN training with portability on a broad variety of computational resources~\cite{Lupo_Pasini_2022}.
HydraGNN is capable of multitask prediction of hybrid node-level and graph-level properties in
large batches of graphs in different sizes, i.e., with varying number of nodes.
We use the ADIOS~\cite{godoy2020adios} high-performance data management library for efficient storage and reading of graph data.
Numerical results show that the training of HydraGNN scales linearly up to 1,024 GPUs on both supercomputers. 

The remainder of this work is structured as follows. Section \ref{section_related} describes related work. Section \ref{sec-background} presents the background of GCNN architectures, the HydraGNN library and the ADIOS data management framework. Section \ref{section_ddp} briefly describes our DDP approach along with efficient data loading of training data. Section \ref{section_numerical_results} presents the numerical results with comparisons between Summit and Perlmutter in terms of scalability, and Section \ref{section_conclusions} summarizes the analysis of this work and discusses future research directions. 
\begin{figure}[t]
\centering
\includegraphics[width=\textwidth]{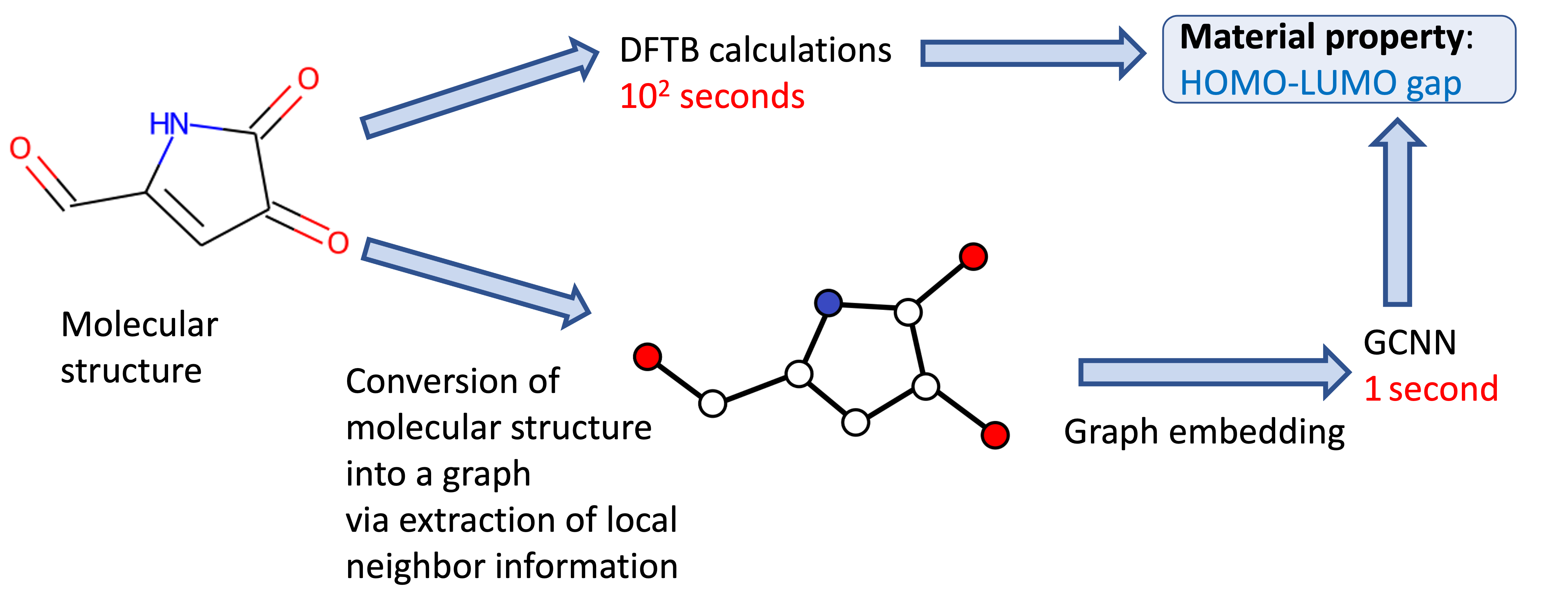}
\caption{ Computational workflow that compares the standard procedure to predict material properties with DFTB calculations and a GCNN model 
that uses the molecular structure as input to estimate the HOMO-LUMO gap. Once the GCNN model is trained, it is much faster than DFTB.}
\label{fig:HydraGNN-High-level-overview}
\end{figure}

\section{Related Work}
\label{section_related}

In this section, we review the relevant literature on GCNN 
models for predicting the HOMO-LUMO gap and DDP for GCNN.

GCNN modeling work has been reported on predictions of HOMO-LUMO gap~\cite{megnet,gilmer2017neural,choudhary2021atomistic,nakamura2020, rahaman2020}. Most of the work focuses on the QM9 dataset~\cite{ramakrishnan2014quantum} or the OE62 dataset~\cite{stuke2020atomic}. 
QM9 has about 134 thousand molecules containing 5 element types (i.e., H,C,N,O and F), and OE62 covers 16 different elements types (i.e., H, Li, B, C, N, O, F, Si, P, S, Cl, As, Se, Br, Te, and I) and about 62 thousand molecules.
The two datasets used in this work are significantly larger and more diverse and hence more challenging to process and predict---PCQM4Mv2 with 3.3 million molecules and 31 element types and AISD  HOMO-LUMO with 10.5 million molecules and 6 element types. Some GCNN work~\cite{hu2021ogb, NEURIPS2021_f1c15925, park2022grpe} has been reported on PCQM4Mv2, but the implementations are not for large-scale distributed training.  


With respect to distributed training of GCNN models, recent work surveyed different types of potential parallelization techniques, including DDP~\cite{https://doi.org/10.48550/arxiv.2205.09702}. While most of the research has focused on distributing the GCNN training with graph partitioning techniques to process large-scale graphs, no specific contribution in the literature has analyzed the scalability of GCNN training using DDP. 
In this paper, we focus on exploring the challenges and demonstrating the capability to process millions of graphs with DDP at scale.


For efficient storage and loading of large datasets, we use the ADIOS~\cite{godoy2020adios} data management library, which is commonly used for science applications. 
ADIOS provides a self-describing data format built upon a publish-subscribe framework.
Other scientific data formats include the Hierarchical Data Format (HDF5)~\cite{hdf5}, but we selected ADIOS for its proven performance at extreme scale, in addition to the plethora of options for tuning I/O and streaming data, along with inbuilt support for data compression.

\section{Background}\label{sec-background}
In this section, we discuss our use of GCNN for the prediction of HOMO-LUMO gap in molecules.
We describe the architecture of HydraGNN, our library of GCNNs.
We discuss the design of the large-scale data loading in HydraGNN leveraging ADIOS, a high-performance scientific data management library for managing large-scale I/O and data transfers.

\subsection{Graph convolutional neural networks}
\label{section_gcnn}

\begin{figure*}[t]
\centering
\includegraphics[width=\textwidth]{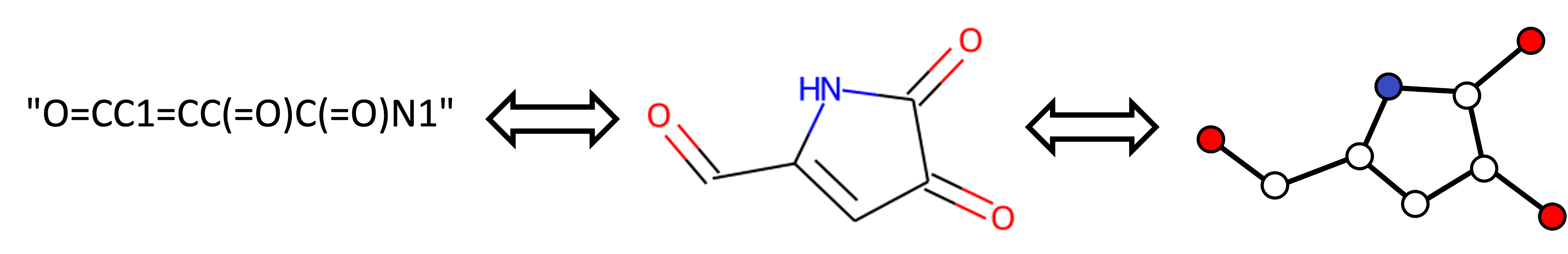}
\caption{Illustration of a SMILES representation of a molecule (left), its corresponding molecular structure (center), and its corresponding molecular graph (right). The atoms that are not explicitly denominated in the molecular structure are carbon atoms and the hydrogen atoms that create a covalence bond with a carbon atom. All the hydrogen atoms are suppressed in the figure for the sake of brevity but they are treated as nodes of the molecular graph in GCNN models.
}
\label{fig:graph-molecule}
\end{figure*}

GCNNs~\cite{GNNpaper, GCNNpaper} are DL models for processing graph data.  
Representing molecules in the form of graphs is natural since the atoms can be viewed as nodes and chemical bonds as edges of the graph, as shown in Figure \ref{fig:graph-molecule}. 
Nodes in the graph retain atomic features in molecules and edges retain the connectivity and bond properties (e.g., the distance between nodes). Each graph can have graph-level properties such as the HOMO-LUMO gap.

Graph convolutional (GC) layers are the core of GCNNs. 
They employ a message-passing framework, a procedure that combines the knowledge from neighboring nodes.
The type of information passed through a graph structure can be either related to the topology of the graph or the nodal features. 
An example of topological information is the node degree, whereas an example of  nodal feature in the context of this work is the atomic number.
The message passing in a single GC layer in our applications maps directly to the pairwise interactions of an atom with its neighbors.
Through consecutive steps of message passing (i.e., stacking multiple GC layers together), the graph nodes gather information from nodes that are further and further away, which implicitly represents many-body interactions.

A graph pooling layer is connected at the end of a stack of consecutive GC layers to gather feature information from the entire graph in prediction tasks of graph-level properties. It aims at aggregating the nodal feature associated with each atom across a graph into a single feature. In our work, we use global mean pooling layer, which averages node features across all the nodes in the graph. 
For atom(node)-level properties such as the atomic charge transfer and atomic magnetic moment, aggregating the information from all atoms into a global feature is not needed.
Finally, fully connected (FC) layers take the results of pooling, i.e., extracted features, and provide the output prediction for global properties.

Differing in the policy adopted to aggregate, transfer, and update information through the message passing in GC layers, a variety of GCNNs have been developed, e.g., Principal Neighborhood Aggregation (PNA)~\cite{corso_principal_2020}, Crystal GCNN (CGCNN)~\cite{cgcnn} and GraphSAGE~\cite{hamilton2017inductive}. Many of them have been implemented in HydraGNN~\cite{hydragnn}.

\subsection{HydraGNN}
\label{subsection_hyragnn}


HydraGNN is 
our in-house library for performant creation and testing of various GCNN models, and is designed to perform multi-task predictions of graph data. It is built on Pytorch~\cite{pytorch, pytorch2019} and Pytorch Geometric~\cite{fey_2019, torch_geometric},  and can run on small-scale workstations to large-scale HPC systems. 
It is openly available on Github~\cite{hydragnn}.

\begin{figure}[t]
    \centering
    \includegraphics[width=0.7\linewidth]{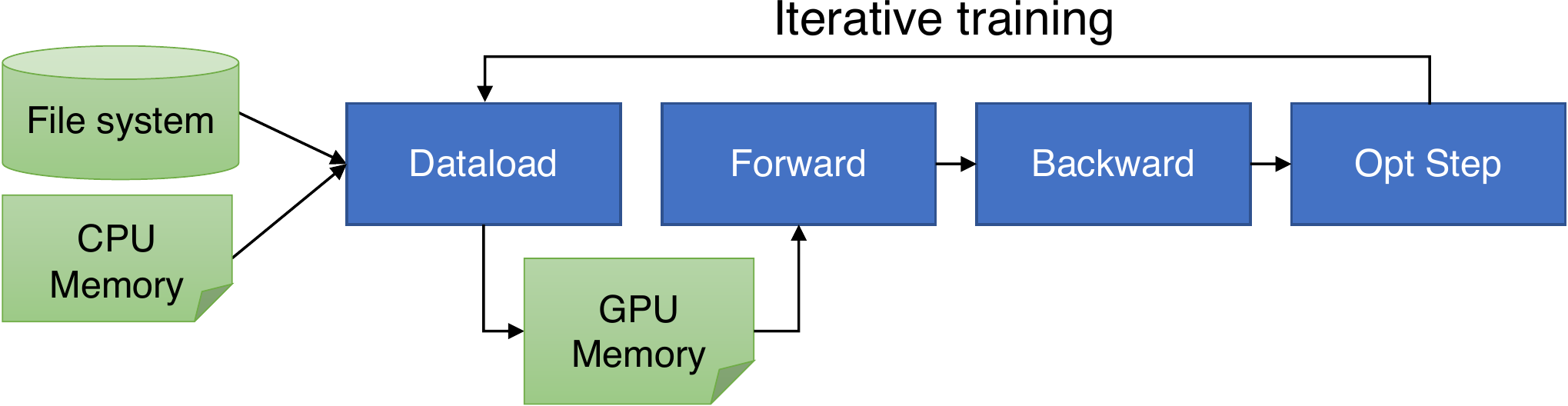}
    \caption{HydraGNN workflow to process molecular graph dataset. It performs iterative training phases, consisting of data loading, forward, backward, and optimization steps. 
    }
    \label{fig:worflow}
\end{figure}

HydraGNN loads molecular data encoded as graph structures consisting of a list of nodes (atoms) and edges (bonds) either from a file system or directly from memory. 
Training is performed over multiple iterations, where each iteration consists of a forward, backward, and optimization step as shown in Figure~\ref{fig:worflow}.
The forward phase computes the model output tensor from an input graph with a forward function consisting of GC layers, a global pooling layer, and FC layers.
It calculates the loss between the model output and the true tensor values corresponding to the input graph as the mean square error (MSE).
The backward phase calculates the gradients of the loss with respect to each parameter in forward function. 
Finally, the optimization step updates parameters based on gradients calculated in the backward step and a user-defined optimization policy.

\subsection{ADIOS Data Management Framework}

ADIOS is an open source, high-performance, I/O framework developed as part of the Exascale Computing Project (ECP)\footnote{https://www.exascaleproject.org}. It provides a custom, self-describing data format with optimized methods to read and write data for massively parallel applications. 
ADIOS's focus is to provide extreme-scale I/O capabilities on the world's largest supercomputers; it is used in science applications that generate data on the order of several petabytes.
ADIOS is run in production in many HPC codes such as XGC~\cite{dominski2021spatial}, GENE~\cite{merlo2021first}, GEM~\cite{cheng2020spatial}, PIConGPU~\cite{poeschel2021transitioning}, WarpX~\cite{wan2021improving}, E3SM~\cite{wang2017data}, LAMMPS~\cite{LAMMPS}, and others, providing over 1 Terabyte/second of I/O to the Summit GPFS file system~\cite{summit} at ORNL.

At its core, ADIOS provides a concept of abstract ``Engine''. Engines execute the I/O heavy tasks and are conceived as workflows tackling specific applications' needs optimized for HPC filesystems, hierachical data management, or data access over the wide area network. We leverage ADIOS's optimized I/O writing and reading performance on HPC file systems in the presence of multiple concurrent processes, which is necessary for performing large-scale distributed data-parallel training for HydraGNN.

Besides the performance boost, ADIOS provides a scalable yet flexible data format that can manage millions of graphs with varying sizes.
ADIOS stores data in a custom format termed BP (binary-packed).
It is a highly efficient self-describing format for storing data along with metadata from various sources in a distributed application.
An ADIOS file consists of variables and attributes, where variables can be scalars or arrays.
Array variables can be local to a process or can be global arrays that are distributed amongst processes.
Global arrays are used to construct large data structures that can be written and read efficiently in a parallel fashion.
For example, edge attributes of all graphs in HydraGNN are stored as part of a single global array for fast storage and retrieval.

Under the hood, an ADIOS file is a container of one or more subfiles. ADIOS allows users to control the number of subfiles depending on the performance needs. A single subfile can be easy to manage but can cause contention when writing or reading in parallel. On the other hand, too many subfiles can overload the metadata server of the filesystem. 
On an HPC system, this is a critical feature. Naive storage approaches that use a separate file for each graph can easily lead to several millions to billions of files for large datasets; on the other hand, using ADIOS to control the number of actual files on the filesystem to store large datasets can significantly alleviate the metadata overhead.

\section{Distributed Data Parallel Training}
\label{section_ddp}

Large supercomputers available at DOE national laboratories such as Summit and Perlmutter allow us to process millions of molecules (graph objects) concurrently to build and test various GCNN models.
To explore the high degree of parallelism available on these nodes, we apply DDP to accelerate the training process in HydraGNN. 



\begin{figure}[t]
    \centering
    \includegraphics[width=0.7\linewidth]{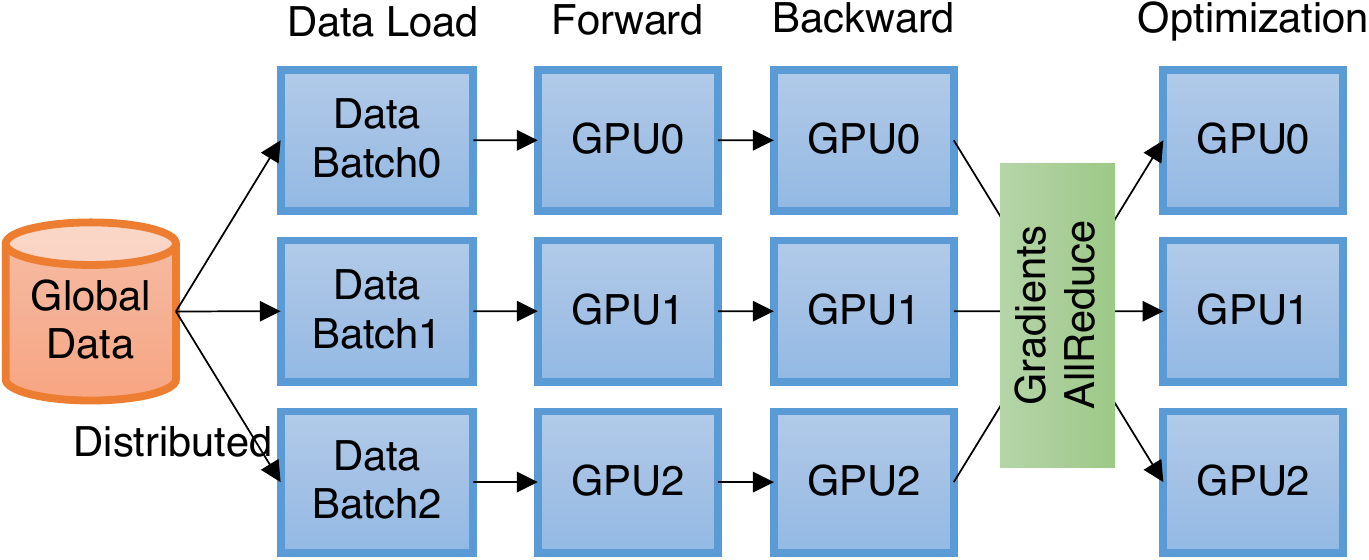}
    \caption{Overview of DDP training. Each process independently loads and processes a batch of data and synchronizes local gradients with others through a gradient aggregation process which requires global communications.}
    \label{fig:ddp}
\end{figure}


In deep learning, DDP refers to a method of performing training in parallel by distributing the same model across multiple nodes but assigning disjoint datasets to consume for each model. In other words, each process computes gradients of parameters in parallel for its assigned dataset. At the end of the computation, we aggregate gradients and re-distribute them to all processes to make each process maintain the same model parameters (Figure~\ref{fig:ddp}). In general, gradients aggregation is an expensive operation in parallel computing. For Summit and Perlmutter, both equipped with NVIDIA GPUs, we use NCCL, an NVIDIA library to communicate directly between GPUs, without having to copy data to the host CPU first, which results in an efficient and inexpensive inter-process communication for gradient aggregation.
The communication overhead will vary depending on the models' size and the frequency of aggregation. 

Another challenge in DDP training in an HPC system is data management, including pre-processing and data loading during training. Each process regularly reads a group of data objects from the storage to form a batch of graph objects for training. 
The random I/O access pattern is common to maintain shuffled and disjointed subsets with other processes to avoid duplication during training and increase the fairness of training.
However, without careful data management, loading training data can become a bottleneck for training. To mitigate this issue, we leverage ADIOS~\cite{godoy2020adios} for our distributed GCNN training. 

We emphasize here the importance of pre-processing.
Many open-source molecular databases use the simplified molecular-input line-entry system (SMILES)~\cite{Weininger1988}, the de facto standard format to represent 2D molecular structures in text strings (see Figure~\ref{fig:graph-molecule} for an example). HydraGNN employs a pre-processing step to convert SMILES strings to graph representations in files. Depending on the scale of the input data (graph), an optimal pre-processing step for graph conversion is crucial to the performance. Without pre-processing, data loading has to perform conversion of SMILES data into a graph which can cause a bottleneck in training. 
To avoid this, our workflow includes a parallel pre-processing step in which graph data generated from SMILES strings is stored into the ADIOS high-performance data format, and is later read back for training. 
This avoids converting the SMILES data into graph objects for every iteration of the training.

To this end, we develop the ADIOS schema for graph datasets. The schema describes the mapping of graph objects to ADIOS components. We aggregate each graph attribute into a global multi-dimensional array and save it in ADIOS variable format. 
The main graph attributes are node features, edge attributes, and edge index, as summarized in 
Table~\ref{tab:schema}.

\begin{table}[t]
\caption{ADIOS schema for graph dataset.}
\label{tab:schema}
\begin{center}
\begin{minipage}{\textwidth}
\begin{tabular*}{\textwidth}{@{}lll@{}}
\toprule
Variable & Description & Array Shape \\
\midrule
x & Features associated to nodes & (\#nodes, \#node features) \\
edge index & Edge connectivity between nodes & (\#2, \#edges) \\
edge attr & Features associated to edges & (\#edges, \#edge features) \\
y & Target features in graph-level or node-level & (\#target, \#features) \\
\botrule
\end{tabular*}
\end{minipage}
\end{center}
\end{table}




We have developed an extensible data loader module in HydraGNN that allows reading data from different storage formats.
In this work, we evaluate the following three methods for loading data from training datasets.
\begin{itemize}
\item Inline data loading: load SMILES strings written in CSV format into memory and then convert each SMILES into a graph object at every batch data loading. It has the smallest memory footprint.
\item Object data loading: convert all SMILES strings into graph objects and export them in a serialized format (e.g., Pickle) during a preprocessing phase. A process loads each data batch directly from the file system and unpacks into a memory.
\item ADIOS data loading: convert SMILES strings into graph objects mapped to ADIOS variables, and write them into an ADIOS file in a pre-processing step. Each process then loads its batch data during training in parallel along with other processes. 
\end{itemize}
We will discuss performance comparisons in the next section.

\section{Numerical Results}
\label{section_numerical_results}

In this section, we assess our development of DDP training in HydraGNN on two state-of-the-art DOE supercomputers, Summit and Perlmutter.
We discuss the scalability of our approach, and compare the performance of different I/O backends for storing and reading graph data.

\subsection{Setup}
We perform our evaluation on two supercomputers of DOE. Both systems provide state-of-the-art GPU-based heterogeneous architectures.

\uline{Summit} is a supercomputer at OLCF, one of DOE's Leadership Computing Facilities (LCFs). 
Summit consists of about 4,600 compute nodes. Each node has a hybrid architecture containing two IBM POWER9 CPUs and six NVIDIA Volta GPUs connected with NVIDIA’s high-speed NVLink. Each node contains 512 GB of DDR4 memory for CPUs and 96 GB of High Bandwidth Memory (HBM2) for GPUs. Summit nodes are connected in a non-blocking fat-tree topology using a dual-rail Mellanox EDR InfiniBand interconnection.

\uline{Perlmutter} is a supercomputer at NERSC. 
Perlmutter consists of about 3,000 CPU-only nodes and 1,500 GPU-accelerated nodes. We use only the GPU-accelerated nodes in our work. Each GPU-accelerated node has an AMD EPYC 7763 (Milan) processor and four NVIDIA Ampere A100 GPUs connected to each other with NVLink-3. Each GPU node has 256 GB of DDR4 memory and 40 GB HBM2 per each GPU.
All nodes in Perlmutter are connected with the HPE Cray Slingshot interconnect. 


We demonstrate the performance of HydraGNN using two large-scale datasets, a previously published benchmark dataset for graph-based learning (PCQM4Mv2)~\cite{Nakata2017,Hu2020} and a custom dataset generated for this work (AISD HOMO-LUMO)~\cite{Blanchard2022AISD}.
For both datasets, molecule information is provided as SMILES strings.
The PCQM4Mv2 
consists of HOMO-LUMO gap values for about 3.3 million molecules. In total, 31 different types of atoms (i.e., H, B, C, N, O, F, Si, P, S, Cl, Ca,Ge,As, Se, Br, I, Mg, Ti, Ga, Zn, Ar, Be, He, Al, Kr, V, Na, Li, Cu, Ne, Ni) are involved in the dataset.
The custom AISD HOMO-LUMO dataset was generated using molecular structures from previous work~\cite{Blanchard2021GB}. 
It is a collection of approximately 10.5 million molecules and contains 6 element types (i.e., H, C, N, O, S, and F).

\begin{table}[t]
\footnotesize
\caption{Dataset description.}
\label{tab:dataset}
\begin{center}
\begin{minipage}{\textwidth}
\begin{tabular*}{\textwidth}{@{\extracolsep{\fill}}lccccccc@{\extracolsep{\fill}}}
\toprule%
& \multicolumn{4}{@{}c@{}}{Graph Size} & \multicolumn{3}{@{}c@{}}{Data File Size (GB)}\\
\cmidrule{2-5}\cmidrule{6-8}%
Dataset & \#Graphs & \#Nodes & \#Edges & \#Avg\footnotemark[1] & CSV   & Pickle & ADIOS \\
\midrule
PCQM4Mv2       & 3.6 M & 105.8 M & 214.6 M & 29.4 & 0.16 & 34 & 22 \\
AISD HOMO-LUMO & 10.5 M & 550.6 M & 1.1 B & 52.4 & 0.88 & 94 & 60 \\
\botrule
\end{tabular*}
\footnotetext[1]{Average number of nodes (atoms) per graph}
\end{minipage}
\end{center}
\end{table}


For scalability tests, we use HydraGNN with 6 PNA~\cite{corso_principal_2020} convolutional layers and 55 neurons per PNA layer. 
The model is trained by using the AdamW method~\cite{DBLP:conf/iclr/LoshchilovH19} with a learning rate of 0.001, local batch size of 128, and maximum epochs set to 3. 
The training set for each of the NN represents 94\% of the total dataset; the validation and test sets each represent $1/3^{rd}$ and $2/3^{rd}$ parts respectively of the remaining data.
For the error convergence tests, the HydraGNN model uses 200 neurons per layer.

\subsection{Scalability of DDP}

\begin{figure}[t]
    \centering
    \includegraphics[width=\linewidth]{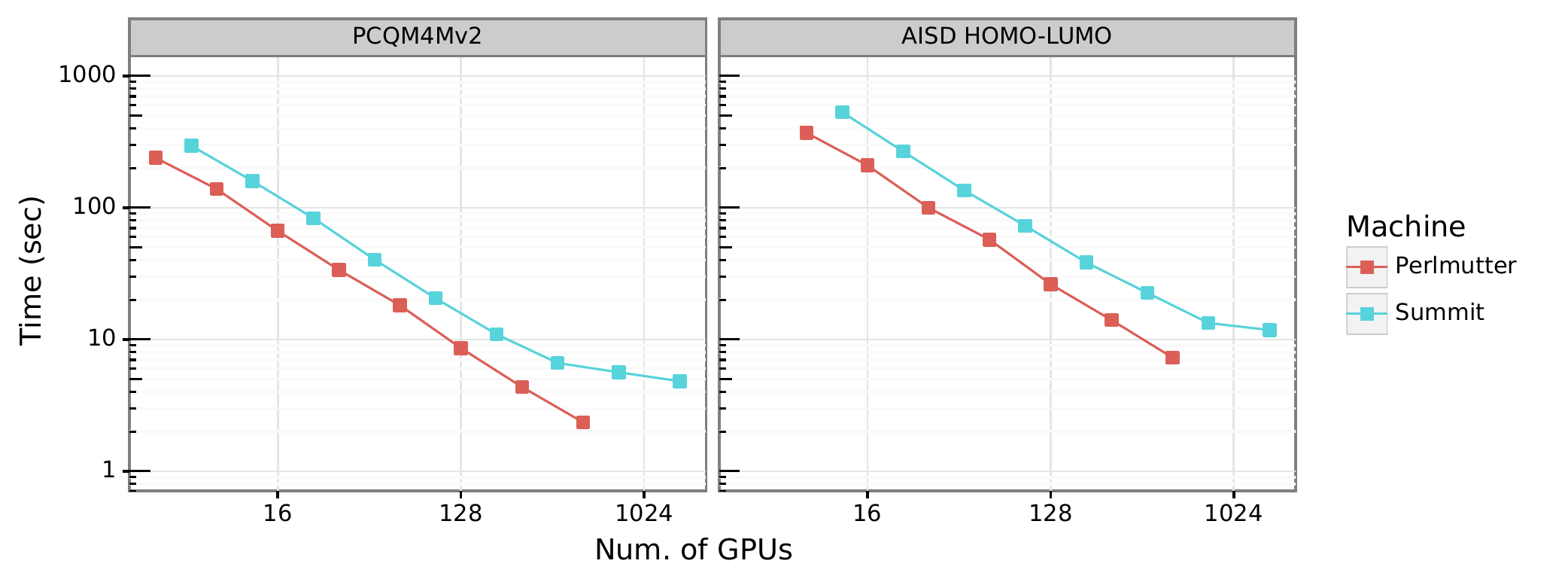}
    \includegraphics[width=\linewidth]{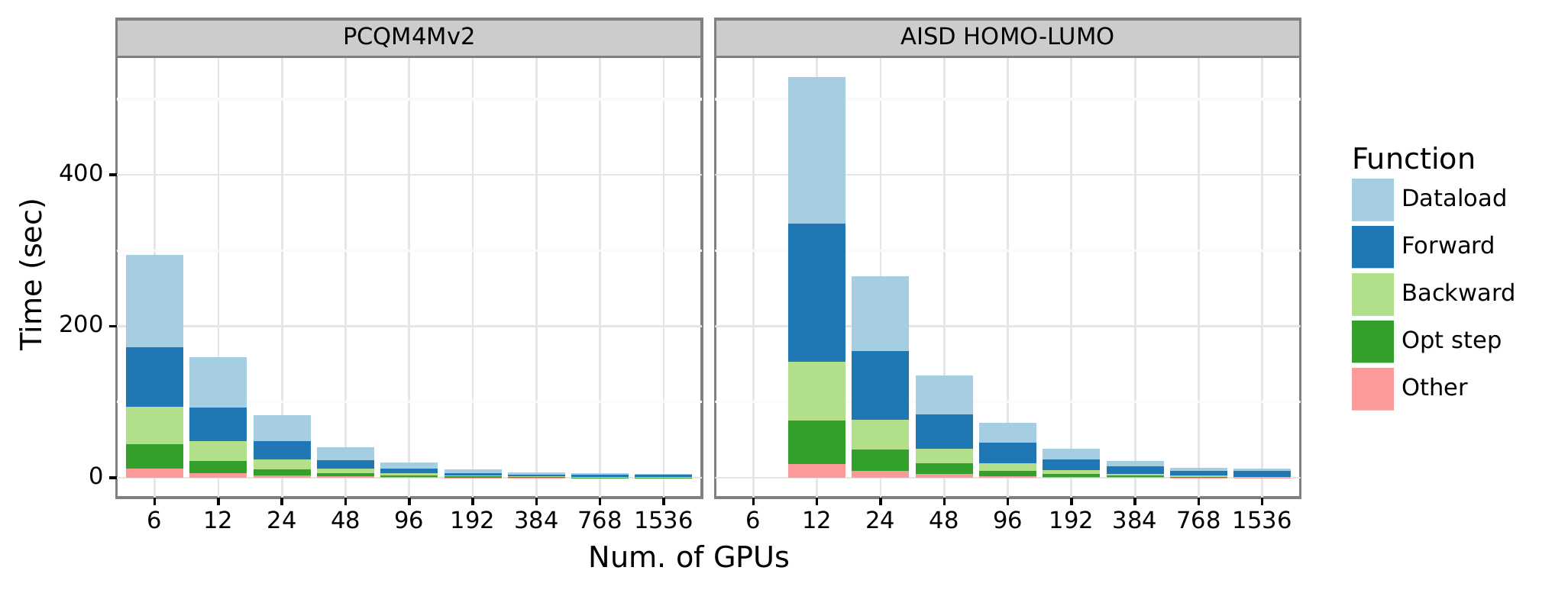}
    \caption{Strong scaling performance of HydraGNN training on OLCF's Summit and NERSC's Perlmutter(top), and detailed timing (bottom).  We perform data-parallel training for PCQM4Mv2 and AISD HOMO-LUMO data sets with  HydraGNN using up to 1,500 GPUs and observe linear scaling up to 1,024 GPUs.}
    \label{fig:scaling}
\end{figure}

We perform DDP training with HydraGNN for PCQM4Mv2 and AISD data on Summit at ORNL and Perlmutter at NERSC using multiple CPUs and GPUs. The number of graphs and size in the datasets are summarized in Table~\ref{tab:dataset}.

We measure the total training time for PCQM4Mv2 and AISD HOMO-LUMO datasets over three epochs. As discussed previously, each training consists of a data loading phase, followed by forward calculation, backward calculation, and optimizer update. We test the scalability of DDP by varying the number of nodes on each system, ranging from a single node up to 256 nodes on Summit and 128 nodes on Perlmutter, corresponding to using 1,536 Volta GPUs and 512 A100 GPUs respectively. 
Figure~\ref{fig:scaling} shows the result. The scaling plot (top) shows the averaged training time for PCQM4Mv2 and AISD HOMO-LUMO on each system with a varying number of nodes, and the detailed timings of each sub-function during the training on Summit are shown at the bottom.

We obtain near-linear scaling up to 1,024 GPUs for both PCQM4Mv2 and AISD HOMO-LUMO data. 
As we further scale the workflow on Summit, the number of batches \textit{per GPU} decreases, leading to sub-optimal utilization of GPU resources. 
As a result, we see a drop in speedup as we scale beyond 1,024 GPUs on Summit. 
We expect similar scaling behavior on Perlmutter, but we were limited to using 128 nodes (i.e., 512 GPUs) for this work.

\subsection{Comparing Different I/O Backends}
\begin{figure}[t]
    \centering
    \includegraphics[width=0.7\linewidth]{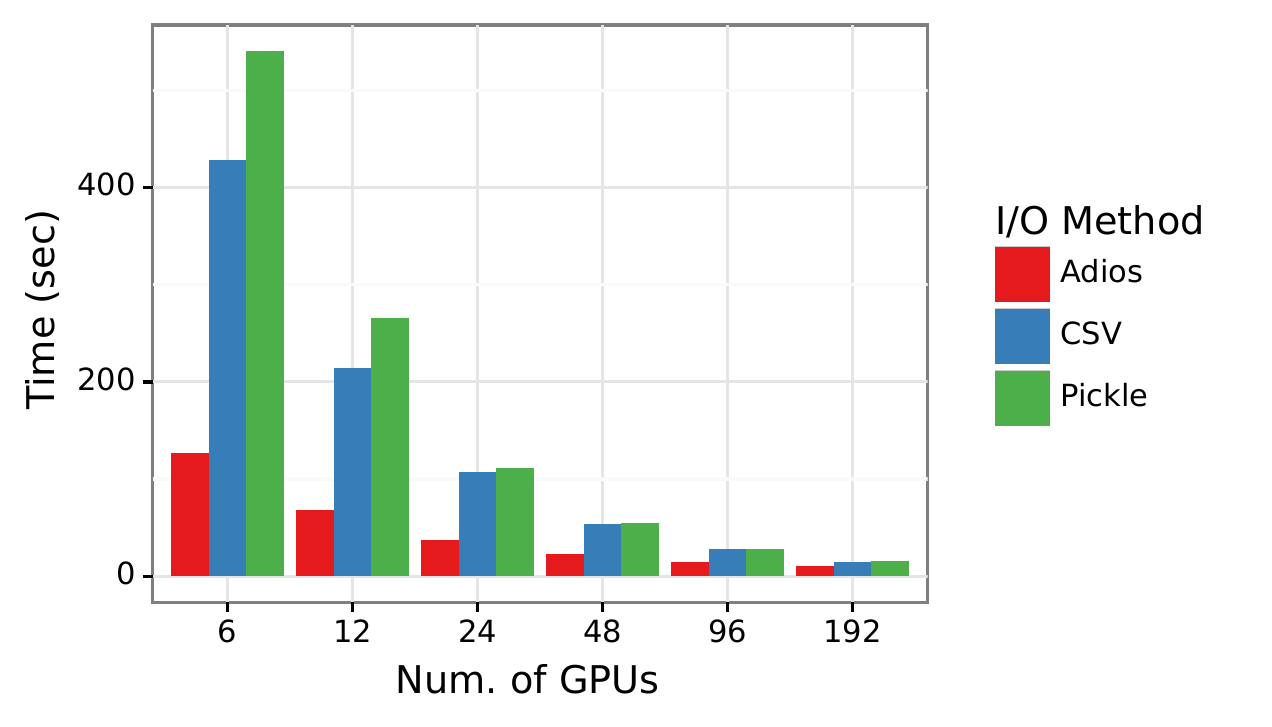}
\caption{Comparison of different I/O methods in HydraGNN. We measure ADIOS data loading time compared with CSV and Pickle with PCQM4Mv2 dataset on Summit.}
    \label{fig:scaling-io}
\end{figure}

Data loading takes a significant amount of time in training, as shown in Figure \ref{fig:scaling}, and hence is a crucial step in the overall workflow.
We compare three different data loading methods -- inline, object loading, and ADIOS data loading, as discussed in~Section~\ref{section_ddp}. 
Figure~\ref{fig:scaling-io} presents time taken by the three methods for the PCQM4Mv2 data set on Summit. 
As expected, it outperforms the CSV data loading test case in which SMILES data is converted into a graph object for every molecule. 
ADIOS outperforms Pickle-based data loading by 4.2x on a single Summit node and 1.5x on 32 Summit nodes.

\subsection{Accuracy}

Next, we perform long-running HydraGNN training for the HOMO-LUMO gap prediction with PCQM4Mv2 and AISD HOMO-LUMO datasets until training converges.

\begin{figure} [t]
    \centering
    \includegraphics[width=0.75\linewidth]{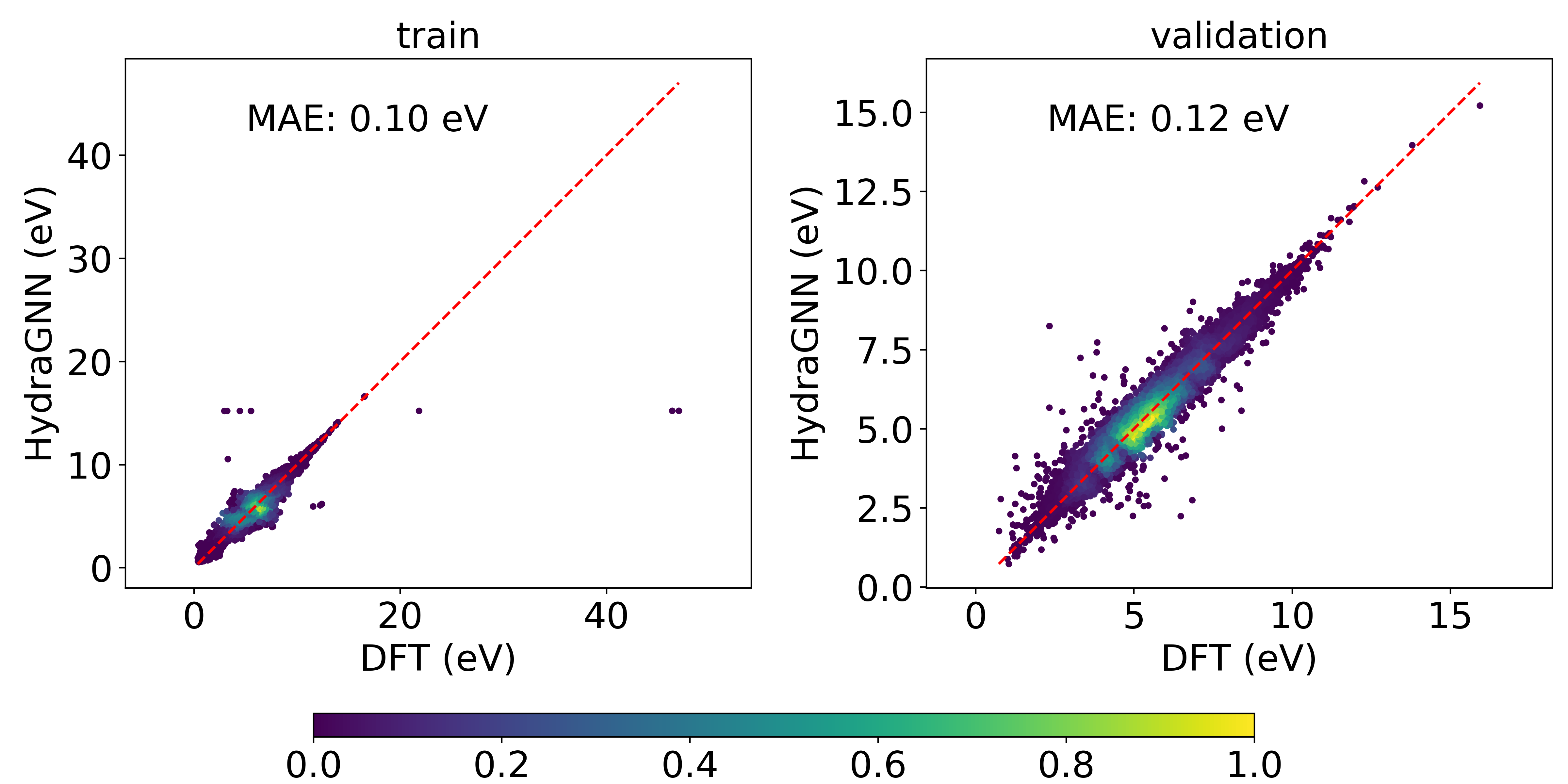}
\caption{HydraGNN predicted values against DFT values of HOMO-LUMO Gap for molecules in PCQM4Mv2 training and validation sets.}
    \label{fig:accuracy_ogb}
\end{figure}

\begin{figure}[t]
    \centering
    \includegraphics[width=\linewidth]{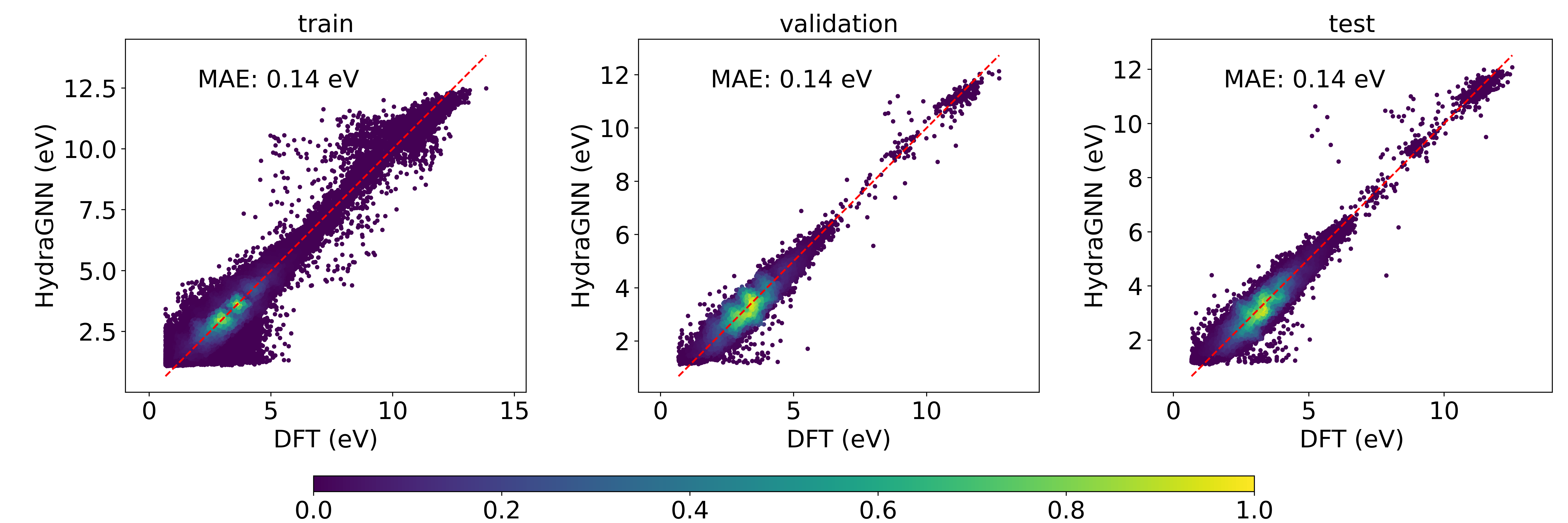}
\caption{HydraGNN predicted values against DFT values of HOMO-LUMO Gap for molecules in AISD HOMO-LUMO training, validation and test sets.}
    \label{fig:accuracy_csce}
\end{figure}

\begin{figure}[t]
    \centering
    \includegraphics[width=0.7\linewidth]{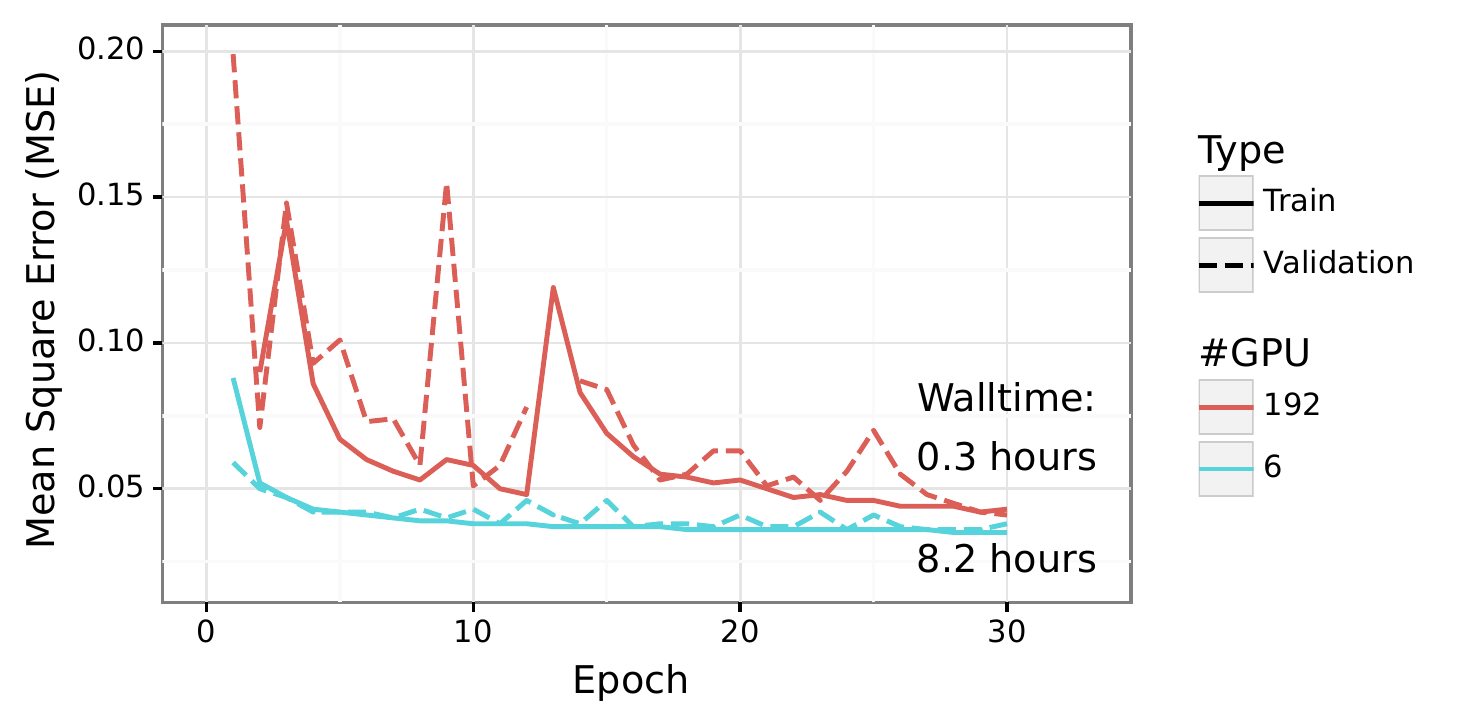}
    \caption{Convergence of the training and validation runs for the AISD HOMO-LUMO data on Summit with different GPU counts. 
    }
    \label{fig:csce-long}
\end{figure}

Figures~\ref{fig:accuracy_ogb} and~\ref{fig:accuracy_csce} show the prediction results for PCQM4Mv2 and AISD HOMO-LUMO datasets respectively. With PCQM4Mv2, we achieve a prediction error of around 0.10 and 0.12 eV, measured in mean absolute error (MAE), for the training and validation set, respectively. 
We note that PCQM4Mv2 is a public dataset released without test data for the purpose of maintaining the OGB-LSC Leaderboards\footnote{{https://ogb.stanford.edu/docs/lsc/leaderboards/}}.
The reported validation MAE from multiple models varies from 0.0857 to 0.1760 eV on the Leaderboard. The validation error of 0.12 eV in this work is within the accepted range.
As for the AISD HOMO-LUMO dataset, it contains almost thrice as many molecules as the PCQM4Mv2 dataset. 
The MAE errors for training, validation, and test sets are 0.14 eV, which is similar to the PCQM4Mv2 dataset. 
Figure \ref{fig:csce-long} shows the accuracy convergence on the AISD HOMO-LUMO dataset using different numbers of Summit GPUs. It shows that the HydraGNN training with 192 GPUs quickly converged in 0.3 hours (wall time) to the similar accuracy level achieved by the 6 GPUs (a single Summit node) that took about 8.2 hours.

We highlight that the convergence of the distributed GCNN training with 192 GPUs is deteriorated compared to the distributed training with only 6 GPUs. 
This is due to a  well known numerical artifact that destabilizes the training of DL models at large scale and causes a performance drop because large scale DDP training is mathematically equivalent to large-batch training. 
In fact, processing data in large batches significantly reduces the stochastic oscillations of the stochastic optimizer used for DL training, thus making the DL training more likely to be trapped in steep local minima, which adversely affect generalization.
Although the final accuracy of the GCNN training with 192 GPUs is slightly worse than the one obtained using 6 GPUs for training, we emphasize the significant advantage that HPC resources provide in speeding-up the training.  
Better accuracy can be obtained when training DL models at large scale by adaptively tuning the learning rate~\cite{You2017LargeBT, 10.1145/3295500.3356137} or by applying quasi-Newton accelerations~\cite{9763953}, but this goes beyond the focus of our current work. 

\section{Conclusions and Future Work}
\label{section_conclusions}
In this paper, we present a computational workflow that performs DDP training to predict the HOMO-LUMO gap of molecules.
We have implemented DDP in HydraGNN, a GCNN library developed at ORNL,
which can utilize heterogeneous computing resources including CPUs and GPUs. 
For efficient storage and loading of large molecular data, we use the ADIOS high-performance data management framework. 
ADIOS helps reduce the storage footprint of large-scale graph structures as compared with commonly used methods, and provides an easy way to efficiently load data and distribute them amongst processes.
We have conducted 
studies using two molecular datasets on the OLCF's Summit and NERSC's Perlmutter supercomputers.
Our results show the near-linear scaling of HydraGNN for the test datasets up to 1024 GPUs.
Additionally, we present the accuracy and convergence behavior of the distributed training with increasing number of GPUs.

Through efficiently managing large-scale datasets and training in parallel, HydraGNN provides an effective surrogate model for accurate and rapid screening of large chemical spaces for molecular design. 
Future work will be dedicated to integrating the scalable DDP training of HydraGNN in a computational workflow to perform molecular design. 



\section*{Acknowledgements}
Massimiliano Lupo Pasini thanks Dr. Vladimir Protopopescu for his valuable feedback in the preparation of this manuscript.

This work was supported in part by the Office of Science of the Department of Energy and by the Laboratory Directed Research and Development (LDRD) Program of Oak Ridge National Laboratory. 
This research is sponsored by the Artificial Intelligence Initiative as part of the Laboratory Directed Research and Development (LDRD) Program of Oak Ridge National Laboratory, managed by UT-Battelle, LLC, for the US Department of Energy under contract DE-AC05-00OR22725.
An award of computer time was provided by the OLCF Director's Discretion Project program using OLCF awards CSC457 and MAT250 and the INCITE program.
This work used resources of the Oak Ridge Leadership Computing Facility and of the Edge Computing program at the Oak Ridge National Laboratory, which is supported by the Office of Science of the U.S. Department of Energy under Contract No. DE-AC05-00OR22725. 
This research used resources of the National Energy Research Scientific Computing Center (NERSC), a U.S. Department of Energy Office of Science User Facility located at Lawrence Berkeley National Laboratory, operated under Contract No. DE-AC02-05CH11231 using NERSC award ASCR-ERCAP-m4133.



%
\section*{Declarations}
The authors declare that they have no conflict of interest.
\\[3mm]
The AISD HOMO-LUMO dataset has been generated and analysed for this work. It is open-source and accessible at the following URL \url{https://doi.ccs.ornl.gov/ui/doi/394}.

\bibliography{sn-bibliography}

\end{document}